\pdfoutput=1

\documentclass[11pt]{article}

\usepackage{ACL2023}

\usepackage{times}
\usepackage{latexsym}

\usepackage[T1]{fontenc}

\usepackage[utf8]{inputenc}

\usepackage{microtype}

\usepackage{inconsolata}

\usepackage{CJKutf8}

\newcommand{\hz}[1]{\begin{CJK*}{UTF8}{gbsn}#1\end{CJK*}}

\usepackage{hyperref}       
\usepackage{url}            
\usepackage{booktabs}       
\usepackage{nicefrac}       
\usepackage{xcolor}         
\usepackage{multirow}
\usepackage{bm}

\usepackage{times}
\usepackage{latexsym}

\usepackage{amsmath}

\usepackage{algorithm}
\usepackage{algpseudocode}
\usepackage{graphicx}  
\usepackage{CJK}
\usepackage{bbm}
\usepackage{float}
\usepackage{xcolor,colortbl}

\usepackage{bigstrut,bigdelim}
\usepackage{paralist}
\usepackage{diagbox}
\usepackage{wrapfig}
\usepackage{verbatim}
\usepackage{enumitem}

\usepackage{subcaption}
\usepackage{multicol}
\usepackage{multirow}
\usepackage{amssymb}
\usepackage{xcolor}
\usepackage{caption}
\usepackage{cleveref}
\usepackage{acronym}
\usepackage{xspace}
\usepackage{pifont}
\usepackage{lipsum}

\makeatletter
\DeclareRobustCommand\onedot{\futurelet\@let@token\@onedot}
\def\@onedot{\ifx\@let@token.\else.\null\fi\xspace}
\def\eg{\emph{e.g}\onedot} 

\def\ie{\emph{i.e}\onedot}

\def\etc{\emph{etc}\onedot} 
\def\vs{\emph{vs}\onedot}

\makeatother

\makeatletter
\newcommand{\thickhline}{%
    \noalign {\ifnum 0=`}\fi \hrule height 1pt
    \futurelet \reserved@a \@xhline
}
\makeatother

\DeclareMathOperator*{\softmax}{Softmax}

\newcommand{\pcr}[1]{{\fontfamily{pcr}\selectfont #1}}
\acrodef{nlp}[NLP]{natural language processing}
\acrodef{vqa}[VQA]{Visual Question Answering}
\acrodef{roi}[RoI]{Region-of-Interest}
\acrodef{plm}[PLM]{pre-trained language model}
\acrodef{ner}[NER]{named entity recognition}
\acrodef{cnn}[CNN]{convolutional neural network}
\acrodef{nlu}[NLU]{natural language understanding}
\acrodef{mrc}[MRC]{machine reading comprehension}
\acrodef{wwm}[WWM]{whole word masking}
\acrodef{sop}[SOP]{sentence ordering prediction}
\acrodef{mem}[MEM]{Masked Entry Modeling}
\acrodef{llm}[LLM]{large language model}
\acrodef{lm}[LM]{language model}

\Crefname{section}{Sec.}{Sec.}
\Crefname{equation}{Eqn.}{Eqns..}

\title{Shu\={o} W\'{e}n Ji\v{e} Z\`{i}: \\ Rethinking Dictionaries and Glyphs for Chinese Language Pre-training}

    \makeatletter
\def\@fnsymbol#1{\ensuremath{\ifcase#1\or \dagger\or \ddagger\or
   \mathsection\or \mathparagraph\or \|\or **\or \dagger\dagger
   \or \ddagger\ddagger \else\@ctrerr\fi}}
    \makeatother

  \author{
  Yuxuan Wang$^{1,2,3}$, Jianghui Wang$^2$, Dongyan Zhao$^{1,2,3,4}$\Thanks{~~Corresponding author: Dongyan Zhao, Zilong Zheng.} , Zilong Zheng$^{2,4\dagger}$ \\
$^1$ Wangxuan Institute of Computer Technology, Peking University \\
$^2$ Beijing Institute for General Artificial Intelligence (BIGAI) \\
$^3$ Center for Data Science, AAIS, Peking University \\
$^4$ National Key Laboratory of General Artificial Intelligence \\
\texttt{ wyx@stu.pku.edu.cn, wangjianghui@bigai.ai, zhaody@pku.edu.cn, zlzheng@bigai.ai} \\ 
\url{https://github.com/patrick-tssn/CDBert}
}

\begin{document}
\maketitle
\begin{abstract}
We introduce \textsc{CDBert}, a new learning paradigm that enhances the semantics understanding ability of the Chinese Pretrained Language Models~(PLMs) with dictionary knowledge and structure of Chinese characters. We name the two core modules of \textsc{CDBert} as Shuowen and Jiezi, where Shuowen refers to the process of retrieving the most appropriate meaning from Chinese dictionaries and Jiezi refers to the process of enhancing characters' glyph representations with structure understanding. To facilitate dictionary understanding, we propose three pre-training tasks, \ie, Masked Entry Modeling, Contrastive Learning for Synonym and Antonym, and Example Learning. We evaluate our method on both modern Chinese understanding benchmark CLUE and ancient Chinese benchmark CCLUE. Moreover, we propose a new polysemy discrimination task PolyMRC based on the collected dictionary of ancient Chinese. Our paradigm demonstrates consistent improvements on previous Chinese PLMs across all tasks. Moreover, our approach yields significant boosting on few-shot setting of ancient Chinese understanding.


\end{abstract}

\section{Introduction}

 \begin{figure}[t!]
     \centering
     \includegraphics[width=\linewidth]{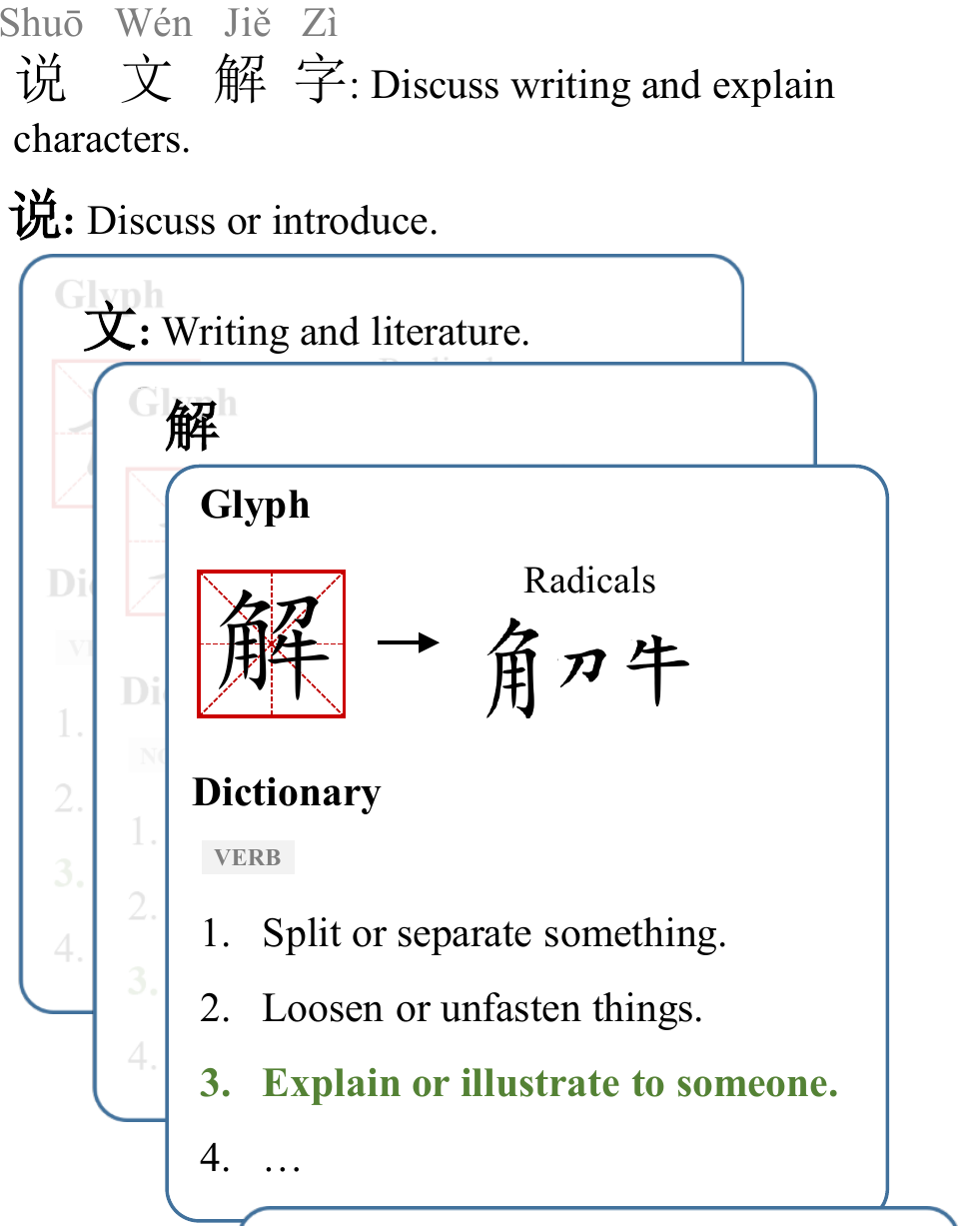}
     \caption{Illustration of \textsc{CDBert}. The expression in green refers to the selected definition of current character.}
     \label{fig:task}
 \end{figure}

Large-scale \acp{plm} such as BERT~\cite{devlin2018bert} and GPT~\cite{brown2020language} have revolutionized various research fields in \ac{nlp} landscape, including language generation~\cite{brown2020language}, text classification~\cite{wang2018glue}, language reasoning~\cite{wei2022chain},  \etc. The \textit{de facto} paradigm to build such LMs is to feed massive training corpus and datasets to a Transformer-based language model with billions of parameters. 

Apart from English \acp{plm}, similar approaches have also been attempted in multilingual~\cite{lample2019cross} and Chinese language understanding tasks~\cite{sun-etal-2021-chinesebert,sun2019ernie}. 
To enhance Chinese character representations,
pioneer works have incorporated additional character information, including glyph~(character's geometric shape), \textit{pinyin}~(character's pronunciation), and stroke~(character's writing order)~\cite{sun-etal-2021-chinesebert,meng2019glyce}. Nevertheless, there still exists a huge performance gap between concurrent state-of-the-art~(SOTA) English \acp{plm} and those on Chinese or other non-Latin languages~\cite{cui-etal-2020-revisiting}, which leads us to rethink the central question: \textit{What are the unique aspects of Chinese that are essential to achieve human-level Chinese understanding?} 

With an in-depth investigation of Chinese language understanding, 
this work aims to point out the following crucial challenges that have barely been addressed in previous Chinese \acp{plm}.

\begin{itemize}[leftmargin=*,noitemsep,topsep=0pt]
    \item \textbf{Frequent \vs Rare Characters.} Different from English that enjoys 26 characters to form frequently-used vocabularies (30,522 WordPieces in BERT), the number of frequently-occurred Chinese characters are much smaller (21,128 in Chinese BERT\footnote{\url{https://github.com/ymcui/Chinese-BERT-wwm}}), of which only 3,500 characters are frequently occurred. As of year 2023, over 17 thousand characters have been newly appended to the Chinese character set. Such phenomenon requires models to quickly adapt to rare or even unobserved characters.
    
    \item \textbf{One \vs Many Meanings.} Compared with English expressions, polysemy is more common for Chinese characters, of which most meanings are semantically  distinguished. 
    Similar as character set, the meanings of characters keep changing. For example, the character ``\hz{卷}'' has recently raised a new meaning: ``the involution phenomena caused by peer-pressure''.
    
    \item \textbf{Holistic \vs Compositional Glyphs.} Considering the logographic nature of Chinese characters, the glyph information has been incorporated in previous works. However, most work treat glyph as an independent visual image while neglecting its compositional structure and relationship with character's semantic meanings.
\end{itemize}

 In this work, we propose \textsc{CDBert}, a new Chinese pre-training paradigm that aims to go beyond feature aggregation and resort to mining information from Chinese dictionaries and glyphs' structures, two essential sources that interpret Chinese characters' meaning. We name the two core modules of \textsc{CDBert} as \textbf{Shuowen} and \textbf{Jiezi}, in homage to one of the earliest Chinese dictionary in Han Dynasty. 
 \Cref{fig:task} depicts the overall model. \textbf{Shuowen} refers to the process that finds the most appropriate definition of a character in a Chinese dictionary. Indeed, resorting to dictionaries for Chinese understanding is not unusual even for Chinese Linguistic experts, especially when it comes to ancient Chinese~(\textit{aka.} classical Chinese) understanding. 
 Different from previous works that simply use dictionaries as additional text corpus~\cite{Yu2021DictBERTEL,chen2022dictbert}, we propose a fine-grained definition retrieval framework from Chinese dictionaries. Specifically, we design three types of objectives for dictionary pre-training: Masked Entry Modeling~(MEM) to learn entry representation; Contrastive Learning objective with synonyms and antonyms; Example Leaning~(EL) to distinguish polysemy by example in the dictionary. \textbf{Jiezi} refers to the process of decomposing and understanding the semantic information existing in the glyph information. Such a process grants native Chinese the capability of understanding new characters. In \textsc{CDBert}, we leverage radical embeddings and previous success of CLIP~\cite{yang2022chinese,radford2021learning} model to enhance model's glyph understanding capability.

 
 We evaluate \textsc{CDBert} with extensive experiments and demonstrate consistent improvements of previous baselines on both modern Chinese and ancient Chinese understanding benchmarks. It is worth noting that our method gets significant improvement on CCLUE-MRC task in few-shot setting. Additionally, we construct a new dataset aiming to test models' ability to distinguish polysemy in Chinese. Based on the $BaiduHanyu$, we construct a polysemy machine reading comprehension task (PolyMRC). Given the example and entry, the model needs to choose a proper definition from multiple interpretations of the entry. We believe our benchmark will help the development of Chinese semantics understanding.  

In summary, the contributions of this work are four-fold: 
(i) We propose \textsc{CDBert}, a new learning paradigm for improving PLMs with Chinese dictionary and characters' glyph representation; 
(ii) We derive three pre-training tasks, Masked Entry Modeling, Contrastive Learning for Synonym and Antonym, and Example Learning, for learning a dictionary knowledge base with a polysemy retriever~(\Cref{sec:dic_pretrain});
(iii) We propose a new task PolyMRC, specially designed for benchmarking model's ability on distinguishing polysemy in ancient Chinese. This new task complements existing benchmarks for Chinese semantics understanding~(\Cref{sec:polymrc});
(iv) We systematically evaluate and analyze the \textsc{CDBert} on both modern Chinese and ancient Chinese NLP tasks, and demonstrate improvements across all these tasks among different types of PLMs. In particular, we obtain significant performance boost for few-shot setting in ancient Chinese understanding.




\section{Related Work}
\paragraph{Chinese Language Model}
Chinese characters, different from Latin letters, are generally logograms. At an early stage, \citet{devlin2018bert, liu2019roberta} propose BERT-like language models with character-level masking strategy on Chinese corpus. \citet{Sun2019ERNIEER} take phrase-level and entity-level masking strategies to learn multi-granularity semantics for \ac{plm}. \citet{Cui2019PreTrainingWW} pre-trained transformers by masking all characters within a Chinese word. \citet{Lai2021LatticeBERTLM} learn multi-granularity information with a constructed lattice graph. Recently, \citet{Zhang2020CPMAL, Zeng2021PanGuLA, Su2022WeLMAW} pre-trained billion-scale parameters large language models for Chinese understanding and generation. In addition to improving masking strategies or model size, some researchers probe the semantics from the structure of Chinese characters to enhance the word embedding. Since Chinese characters are composed of radicals, components, and strokes hierarchically, various works~\cite{Sun2014RadicalEnhancedCC, Shi2015RadicalED, Li2015ComponentEnhancedCC, Yin2016MultiGranularityCW, Xu2016ImproveCW, Ma2020LearningCW, Lu2022LearningCW} learn the Chinese word embedding through combining indexed radical embedding or hierarchical graph. Benefiting from the strong representation capability of \acp{cnn}, some researchers try to learn the morphological information directly from the glyph~\cite{Liu2017LearningCC, Zhang2017WhichEI, Dai2017GlyphawareEO, Su2017LearningCW, Tao2019ChineseEV, Wu2019GlyceGF}.  
\citet{Sehanobish2019UsingCG, Xuan2020FGNFG} apply the glyph-embedding to improve the performance of BERT on \ac{ner}. Besides, polysemy is common among Chinese characters, where one character may correspond to different meanings with different pronunciations. Therefore, \citet{Zhang2019LearningCW} use ``pinyin'' to assist modeling in distinguishing Chinese words. \citet{Sun2021ChineseBERTCP} first incorporate glyph and ``pinyin'' of Chinese characters into \ac{plm}, and achieve SOTA performances across a wide range of Chinese NLP tasks. \citet{su-etal-2022-rocbert} pre-trained a robust Chinese BERT with synthesized adversarial contrastive learning examples including semantic, phonetic, and visual features. 

\paragraph{Knowledge Augmented pre-training}
Although PLMs have shown great success on many NLP tasks. There are many limitations on reasoning tasks and domain-specific tasks, where the data of downstream tasks vary from training corpus in distribution. Even for the strongest LLM ChatGPT, which achieves significant performance boost across a wide range of NLP tasks, it is not able to answer questions involving up-to-date knowledge. And it is impossible to train LLMs frequently due to the terrifying costs. As a result, researchers have been dedicated to injecting various types of knowledge into PLM/LLM. According to the types, knowledge in existing methods can be classified to text knowledge \cite{Hu2022ASO} and graph knowledge, where text knowledge can be further divided into linguistic knowledge and non-linguistic knowledge. Specifically, some works took lexical information \cite{Lauscher2019SpecializingUP, Zhou2019LIMITBERTL, Lyu2021LETLK} or syntax tree \cite{Sachan2020DoST, Li2020ImprovingBW, Bai2021SyntaxBERTIP} to enhance the ability of PLMs in  linguistic tasks. For the non-linguistic knowledge, some researchers incorporate general knowledge such as Wikipedia with retrieval methods \cite{Guu2020RetrievalAL,Yao2022KformerKI,Wang2022TrainingDI} to improve the performance on downstream tasks, others use domain-specific corpora \cite{Lee2019BioBERTAP, Beltagy2019SciBERTAP} to transfer the PLMs to corresponding downstream tasks. Compared with text knowledge, a knowledge graph contains more structured information and is better for reasoning. Thus a flourish of work \cite{Liu2019KBERTEL, Yu2020JAKETJP, He2021KLMoKG, Sun2021JointLKJR, Zhang2022GreaseLMGR} designed fusion methods to combine the KG with PLMs.

\paragraph{Dictionary Augmented pre-training}
Considering the heavy-tailed distribution of the pre-training corpus and difficult access to the knowledge graph, some works injected dictionary knowledge into PLMs to alleviate the above problems. \cite{Yu2021DictBERTEL} enhance PLM with rare word definitions from English dictionaries. \citet{chen2022dictbert} pre-trained BERT with English dictionary as a pre-training corpus and adopt an attention-based infusion mechanism for downstream tasks.

\section{\textsc{CDBert}}\label{sec:cdbert}


\subsection{Shuowen: Dictionary as a pre-trained Knowledge}\label{sec:dic_pretrain}
We take three steps while looking up the dictionary as the pre-training tasks: 1) Masked Entry Modeling (MEM). The basic usage of a dictionary is to clarify the meaning of the entry. 2) Contrastive Learning for Synonym and Antonym~(CL4SA). For ambiguous meanings, we always refer to the synonym and antonym for further understanding. 3) Example Learning (EL). We will figure out the accurate meaning through several classical examples. 

\paragraph{\ac{mem}}
Following existing transformer-based language pre-training models \cite{devlin2018bert,liu2019roberta}, we take the \ac{mem} as a pre-training task. 
Specifically, we concatenate the entry (\pcr{<ent>}) with its corresponding meaning or definition (\pcr{<def>}) as input, \ie, \{\pcr{[CLS] <ent> [SEP] <def> [SEP]}\}. Then the \ac{mem} task masks out the \pcr{<ent>} with a \pcr{[MASK]} token, and attempts to recover it. Considering the entry might be composed of multiple characters, we use \ac{wwm}~\cite{cui-etal-2020-revisiting} as the entry masking strategy. The objective of \ac{mem} $\mathcal{L}_{mem}$ is computed as the cross-entropy between the recovered entry and the ground truth.

\paragraph{Contrastive Learning for Synonym and Antonym~(CL4SA)}
Inspired by \citet{yang2022chinese}, we adopt contrastive learning to better support the semantics of the pre-trained representation. We construct positive sample pair $\langle ent, syno \rangle$ with synonyms in the dictionary, and negative sample pair $\langle ent, anto\rangle$ with antonyms in the dictionary. The goal of the CL4SA is to make the positive sample pair closer while pushing the negative sample pair further. Thus we describe the contrastive objective as follows:
\begin{equation*}
    \mathcal{L}_{cl4sa} = -\log \frac{e^{h_{ent} \cdot h_{syno}}}{e^{h_{ent} \cdot h_{syno}} + e^{h_{ent} \cdot h_{anto}}}
\end{equation*}
where $\cdot$ denotes the element-wise product, $h_{ent}$, $h_{syno}$, $h_{anto}$ is the representation of the original entry, the synonym, and the antonym respectively. In practice, we use the hidden states of \pcr{[CLS]} token as the representation of the input $\{$\pcr{[CLS]}\pcr{<ent>}\pcr{[SEP]}\pcr{<def>}\pcr{[SEP]}$\}$.   Since the antonyms in the dictionary are much less than synonyms, we randomly sampled entries from the vocabulary for compensation. To distinguish the sampled entries with the strict antonyms, we set different weights for them. 

\paragraph{Example Learning~(EL)}
Compared with other languages, the phenomenon of polysemy in Chinese is more serious, and most characters or words have more than one meanings or definitions. To better distinguish multiple definitions of an entry in a certain context, we introduce example learning, which attempts to learn the weight of different definitions for a certain example. Specifically, given an entry $ent$, $K$ multiple definitions ${def}_1, \ldots, {def}_K$, and an exemplar phrase $exa_i$ of meaning ${def}_i$, we use $h_{exa}$, the hidden state of the \pcr{[CLS]} token in the example as query $Q$,  and $X=\{h_{m}^i\}_{i=1}^k$, the hidden states of the \pcr{[CLS]} token in the meanings as key $K$. Then the attention score can be computed as: 
\begin{equation}
    Attn_{def} = \softmax \left(\frac{QK^T}{\sqrt{d_k}}\right)
\end{equation}
We use the cross-entropy loss to supervise the meaning of retriever training:
\begin{equation}
    \mathcal{L}_{el} = CrossEntropy(\text{one-hot}(def), Attn_{def})
\end{equation}
where $\text{one-hot}(\cdot)$ is a one-hot vector transition of ground-truth indexes. 

We sum over all the above objectives to obtain the final loss function:
\begin{equation}
    \mathcal{L} = \lambda_1 \mathcal{L}_{mem} + \lambda_2 \mathcal{L}_{cl4sa} + \lambda_3 \mathcal{L}_{el}
    \label{eqn:tot}
\end{equation}
where $\lambda_1$, $\lambda_2$, $\lambda_3$ are three hyper-parameters to balance three tasks.

\subsection{Jiezi: Glyph-enhanced Character Representation}\label{sec:jiezi}
\begin{figure}
    \centering
    \includegraphics[width=\linewidth]{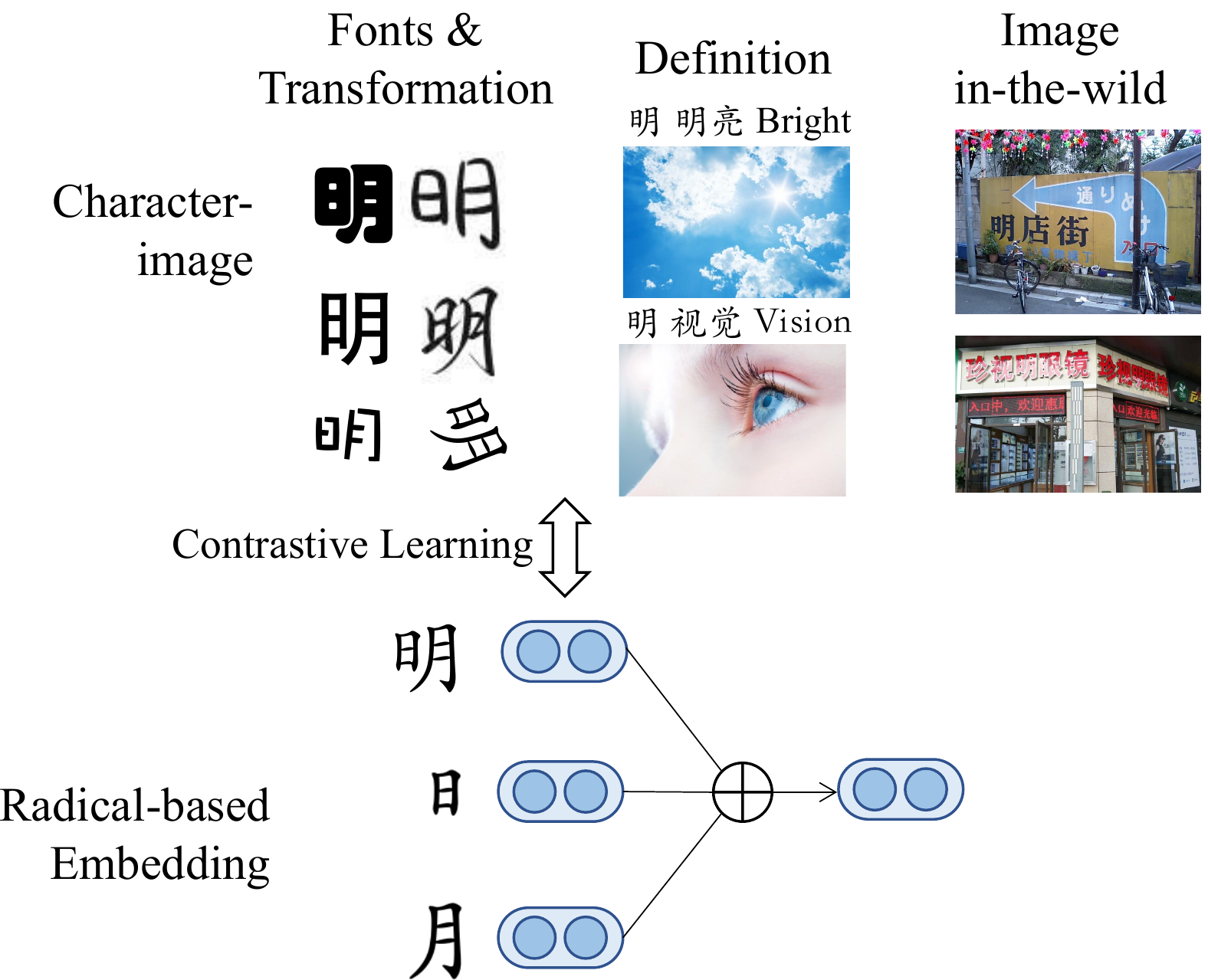}
    \caption{Illustration of glyph-enhanced character representation on character ``\hz{明}''.}
    \label{fig:my_label}
\end{figure}
Chinese characters, different from Latin script, demonstrate strong semantic meanings. We conduct two structured learning strategies to capture the semantics of Chinese characters. Following \citet{sun-etal-2021-chinesebert}, we extract the glyph feature by the CNN-based network. 

\paragraph{CLIP enhanced glyph representation} To better capture the semantics of glyphs, we learn the glyph representation through a contrastive learning algorithm. 
Specifically, we concatenate character $c$ with its definition $def$ as text input and generate a picture of the character as visual input. 
We initialize our model with the pre-trained checkpoint of Chinese-CLIP~\cite{yang2022chinese} and keep the symmetric cross-entropy loss over the similarity scores between text input and visual input as objectives. 
To alleviate the influence of pixel-level noise, we follow \citet{Jaderberg14c,Jaderberg16} to generate a large number of images of characters by transformation, including font, size, direction, \etc. Besides, we introduce some Chinese character images in wild \cite{yuan2019ctw} in the training corpus to improve model robustness. Finally, we extract the glyph feature through the text encoder to mitigate the pixel bias. 

\paragraph{Radical-based character embedding} Since the glyph feature requires extra processing and is constrained by the noise in images, we propose a radical-based embedding for end-to-end pre-training. We first construct a radical vocabulary, then add the radical embedding for each character with their radical token in the radical vocabulary. Thus, we can pre-train the \textsc{CDBert} in the end-to-end learning method.

\subsection{Applying \textsc{CDBert} to downstream tasks}
\begin{figure}[ht!]
    \centering
    \includegraphics[width=\linewidth]{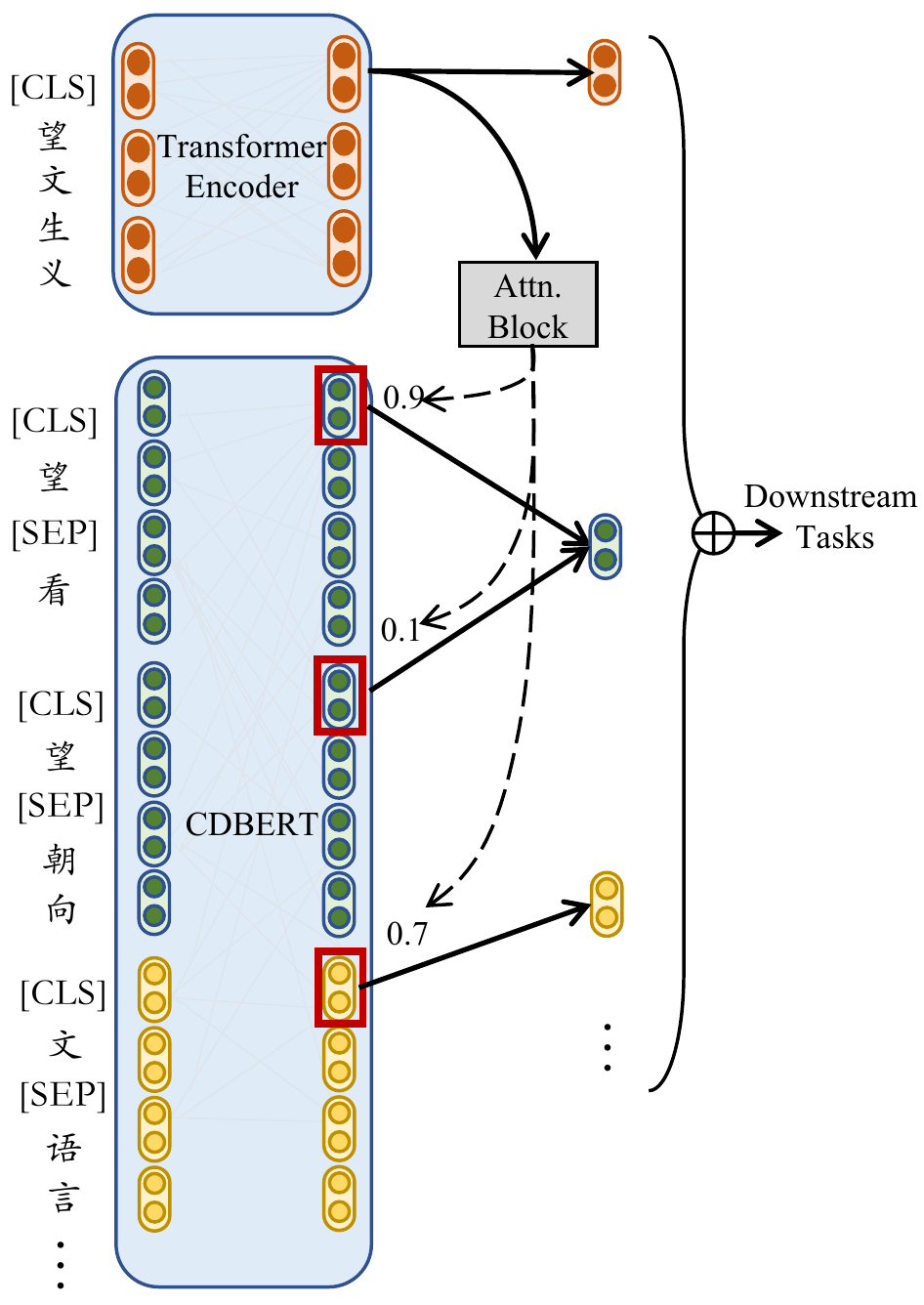}
    \caption{Illustration of applying \textsc{CDBert} to downstream tasks. $\bigoplus$ indicates the concatenation operation. The Attn. Block is the pre-trained attention model from the EL task.}
    \label{fig:my_label}
\end{figure}
Following \citet{chen2022dictbert}, we use the \textsc{CDBert} as a knowledge base for retrieving entry definitions. Specifically, given an input expression, we first look up all the entries in the dictionary. Then, we adopt the dictionary pre-training to get the representation of the entry. At last, we fuse the \textsc{CDBert}-augmented representation to the output of the language model for further processing in downstream tasks. We take the attention block pre-trained by the EL task as a retriever to learn the weight of all the input entries with multiple meanings. After that, we use weighted sum as a pooling strategy to get the \textsc{CDBert}-augmented representation of the input. We concatenate the original output of the language model with the \textsc{CDBert}-augmented representation for final prediction.

\section{The PolyMRC Task}\label{sec:polymrc}
Most existing Chinese language understanding evaluation benchmarks do not require the model to have strong semantics understanding ability. Hence, we propose a new dataset and a new machine reading comprehension task focusing on polysemy understanding. Specifically, we construct a dataset through entries with multiple meanings and examples from dictionaries. As for the Polysemy Machine Reading Comprehension~(PolyMRC) task, we set the example as context and explanations as choices, the goal of PolyMRC is to find the correct explanation of the entry in the example. \Cref{tab:ploymrc_data} shows the statistics of the dataset. 

\begin{table}[ht!]
    \centering
    \caption{Statistics of PolyMRC Dataset}
    \label{tab:ploymrc_data}
    \resizebox{\linewidth}{!}{%
    \begin{tabular}{l|ccccc}
    \thickhline
        \multicolumn{1}{c|}{\textbf{Split}}  & \textbf{Sentences} & \textbf{Average length}    \\ \hline
        Training data & 46,119 & 38.55 \\
        Validation data & 5,765 & 38.31 \\
        Test data & 5,765 & 38.84 \\

         \thickhline
    \end{tabular}
    }
\end{table}

\section{Experiments}

\subsection{Implementation Details}
We pre-train CDBert based on multiple official pre-trained Chinese BERT models. All the models are pre-trained for 10 epochs with batch size 64 and maximum sequence length 256. We adopt AdamW as the optimizer and set the learning rate as $5e-5$ with a warmup ratio of $0.05$. We set $\lambda_1 = 0.6$, $\lambda_2=0.2$, and $\lambda_3=0.2$ in \Cref{eqn:tot} for all the experiments. We finetune CLUE~\cite{xu-etal-2020-clue} with the default setting reported in the CLUE GitHub repository\footnote{\url{https://github.com/CLUEbenchmark/CLUE}}.


\subsection{Baselines}
\paragraph{BERT} We adopt the official BERT-base model pre-trained on the Chinese Wikipedia corpus as baseline models.
\paragraph{RoBERTa} 
Besides BERT, we use two stronger \acp{plm} as baselines:  RoBERTa-base-wwm-ext and RoBERTa-large-wwm-ext (we will use RoBERTa and RoBERTa-large for simplicity). In these models, \textit{wwm} denotes the model continues pre-training on official RoBERTa models with the \ac{wwm} strategy, and \textit{ext} denotes the models are pre-trained on extended data besides Wikipedia corpus. 

\paragraph{MacBERT}
MacBERT improves on RoBERTa by taking the MLM-as-correlation~(MAC) strategy and adding sentence ordering prediction (SOP) as a new pre-training task. We use MacBERT-large as a strong baseline method.

\begin{table*}[ht!]
    \centering
    \caption{Performance improvements of \textsc{CDBert} on $\text{CLUE}_{classification}$.}
    \label{tab:clue_cls}
    \resizebox{\textwidth}{!}{
    \begin{tabular}{l|ccccccc}
    \thickhline
        \multicolumn{1}{c|}{\textbf{Model}}  & \textbf{AFQMC} & \textbf{TNEWS'} &     \textbf{IFLYTEK} & \textbf{CMNLI} & \textbf{WSC} & \textbf{CSL}  & \textbf{SCORE} \\ \hline
         BERT$_{\text{Base}}$ & 73.70 & 56.58 & 60.29 & 79.69 & 70 & 80.36 & 70.10 \\ 
         BERT$_{\text{Base}} + $ \textsc{CDBert}&73.48 & 57.19 & 62.12 & 80.19 & 71.38 & 81.4 & 70.96 \\ 
         RoBERTa$_{\text{ext}}$ & 74.04 & 56.94 & 60.31 & 80.51 & 80.69 & 81 & 72.25\\ 
         RoBERTa$_{\text{ext}} + $ \textsc{CDBert}&74.88 & 57.68 & 62.19 & 81.81 & 81.38 & 80.93 & 73.15 \\ 
         RoBERTa$_{\text{ext-large}}$ & 76.55 & 58.61 & 62.98 & 82.12 & 82.07 & 82.13 & 74.08 \\ 
         RoBERTa$_{\text{ext-large}} + $ \textsc{CDBert}& \textbf{76.82} & \textbf{59.09} & \textbf{63.04} & \textbf{82.89} & \textbf{84.83} & \textbf{83.07} & \textbf{74.95}\\ 
         
         \thickhline
    \end{tabular}
    }
\end{table*}

\begin{table*}[th!]
    \centering
    \caption{Performance of \textsc{CDBert} on CLUE$_{MRC \& QA}$. \small{* we can not reproduce the result reported in CLUE github repo. (<report in github repo>-<report in paper>)}}
    \label{tab:clue_mrc}
    \resizebox{\textwidth}{!}{
    \begin{tabular}{l|cccc}
    \thickhline
        \multicolumn{1}{c|}{\textbf{Model}}  & \textbf{CMRC2018} & \textbf{CHID} &     \textbf{C3} & \textbf{SCORE} \\ \hline
         BERT$_{\text{Base}}$ & 71.6 & 80.04 & 64.50 & 72.71 \\ 
         BERT$_{\text{Base}} + $ \textsc{CDBert}& 71.75 & 82.61 & 65.39 & 73.25 \\ 
         RoBERTa$_{\text{ext}}$ & 75.20 & 83.62 & 66.50 & 75.11 \\ 
         RoBERTa$_{\text{ext}} + $ \textsc{CDBert}& 75.85 & 84.7 & 67.09 & 75.88 \\ 
         RoBERTa*$_{\text{ext-large}}$ & 76.65~(77.95-76.58) & 85.32~(85.37-85.37) & 73.72~(73.82-72.32) & 78.56 \\ 
         RoBERTa$_{\text{ext-large}} + $ \textsc{CDBert}& 77.75 & 85.38 & 73.95 & 79.03 \\ 
         
         \thickhline
    \end{tabular}
    }
\end{table*}

\subsection{CLUE}

We evaluate the general \ac{nlu} capability of our method with CLUE benchmark~\cite{xu-etal-2020-clue}, which includes text classification and \ac{mrc} tasks. There are five datasets for text classification tasks: \textbf{CMNLI} for natural language inference, \textbf{IFLYTEK} for long text classification, \textbf{TNEWS'} for short text classification, \textbf{AFQMC} for semantic similarity, \textbf{CLUEWSC 2020} for coreference resolution, and \textbf{CSL} for keyword recognition. The text classification tasks can further be classified into single-sentence tasks and sentence pair tasks. The \ac{mrc} tasks include span selection-based \textbf{CMRC2018}, multiple choice questions C3, and idiom Cloze ChID.

The results of text classification are shown in \Cref{tab:clue_cls}. In general, \textsc{CDBert} performs better on single-sentence tasks than sentence pair tasks. Specifically, compared with baselines, \textsc{CDBert} achieves an average improvement of $1.8\%$ on single sentence classification: TNEWS', IFLYTEK, and WSC. Besides, \textsc{CDBert} outperforms baselines on long text classification task IFLYTEK by improving $2.08\%$ accuracy on average, which is more significant than the results ($1.07\%$) on short text classification task TNEWS'. This is because TNEWS' consists of news titles in 15 categories, and most titles consist of common words which are easy to understand. But IFLYTEK is a long text 119 classification task that requires comprehensive understanding of the context. In comparison, the average improvement on sentence pair tasks brought by \textsc{CDBert} is $0.76\%$, which is worse than the results on single sentence tasks. These results show dictionary is limited in helping PLM to improve the ability of advanced NLU tasks, such as sentiment entailment, keywords extraction, and natural language inference.

We demonstrate the results on MRC tasks in \Cref{tab:clue_mrc}. As we can see, \textsc{CDBert} yields a performance boost on MRC tasks ($0.79\%$) on average among all the baselines. It is worth noting that when the PLM gets larger in parameters and training corpus, the gain obtained by \textsc{CDBert} becomes less. We believe this is caused by the limitation of CLUE benchmark for the reason that several large language models have passed the performance of humans~\cite{xu-etal-2020-clue}. 

\begin{table*}[th!]
\begin{minipage}{0.63\textwidth}
    \captionof{table}{Performance of \textsc{CDBert} on CCLUE.}
    \label{tab:cclue}
    \resizebox{\linewidth}{!}{
    \begin{tabular}{l|cccccc}
    \thickhline
        \multicolumn{1}{c|}{\textbf{Model}}  & \textbf{NER} &      \textbf{CLS} & \textbf{SENT} & \textbf{MRC} & \textbf{SCORE} \\ \hline
         BERT$_{\text{Base}}$ &  71.62  & 82.31 & 59.95 & 42.76 & 64.16 \\
         BERT$_{\text{Base}} + $ \textsc{CDBert}& 72.41 & 82.74 & 60.25 & 43.91 & 64.83\\ 
         RoBERTa$_{\text{ext}}$ & 69.5 & 81.96 & 59.4 & 42.3 & 63.29 \\ 
         RoBERTa$_{\text{ext}} + $\textsc{CDBert}& 70.89 & 82.15 & 59.95 & 44.14 & 64.28  \\ 
         RoBERTa$_{\text{ext-large}}$ & 79.87 & 82.9 & 58.4 & 43.45 & 66.16 \\ 
         RoBERTa$_{\text{ext-large}} + $ \textsc{CDBert}& 79.93 & 83.03 & 59.75 & 45.52 & 67.06 \\ 
         MacBERT$_{\text{ext-large}}$ & 81.89 & 83.06 & 58.9 & 43.22 & 66.77\\ 
         MacBERT$_{\text{ext-large}} + $ \textsc{CDBert}& 82.33 & 83.71 & 59.4 & 45.29 & 67.68\\ 
         
         \thickhline
    \end{tabular}
    }
\end{minipage}
\hfill
\begin{minipage}{0.36\textwidth}
    \captionof{table}{Performance on PolyMRC.}
    \label{tab:poly_res}
    \resizebox{\linewidth}{!}{%
    \begin{tabular}{l|ccccc}
    \thickhline
        \multicolumn{1}{c|}{\textbf{Model}}  & \textbf{Accuracy}\\ \hline
         BERT$_{\text{Base}}$ & 65.33 \\ 
         BERT$_{\text{Base}} + $ \textsc{CDBert}& 65.93 \\ 
         RoBERTa$_{\text{ext}}$ & 61.96 \\ 
         RoBERTa$_{\text{ext}} + $ \textsc{CDBert}& 62.93 \\ 
         RoBERTa$_{\text{ext-large}}$ & 64.18 \\ 
         RoBERTa$_{\text{ext-large}} + $ \textsc{CDBert}& 64.77 \\ 
         MacBERT$_{\text{ext-large}}$ & 66.73 \\ 
         MacBERT$_{\text{ext-large}} + $ \textsc{CDBert}& 67.16 \\ 
         
         \thickhline
    \end{tabular}
    }
\end{minipage}
\end{table*}
         

\subsection{CCLUE}
Ancient Chinese~(\textit{aka.} Classical Chinese) is the essence of Chinese culture, but there are many differences between ancient Chinese and modern Chinese. CCLUE\footnote{\url{https://cclue.top}} is a general ancient \ac{nlu} evaluation benchmark including \ac{ner} task, short sentence classification task, long sentence classification task, and machine reading comprehension task. 
We use the CCLUE benchmark to evaluate the ability of \textsc{CDBert} to adapt modern Chinese pre-trained models to ancient Chinese understanding tasks.  

In order to assess the ability of modern Chinese PLM to understand ancient Chinese by \textsc{CDBert}, we test our model on CCLUE benchmark. We pre-train the \textsc{CDBert} on the ancient Chinese dictionary for fairness. Results are presented in \Cref{tab:cclue}, which shows \textsc{CDBert} is helpful in all three general NLU tasks: sequence labeling, text classification, and machine reading comprehension. We find in MRC task, \textsc{CDBert} improves from $42.93$ on average accuracy of all 4 models to $44.72$ ($4.15\%$ relatively), which is significantly better than other tasks. In addition, we can see the gain obtained from the model scale is less than \textsc{CDBert} on CLUE datasets. This is because the training corpus of these PLMs do not contain ancient Chinese. In this scenario, \textsc{CDBert} is more robust.

         

\paragraph{PolyMRC Results}
We use BERT, RoBERTa, and MacBERT as baselines for the new task. Considering the context of PolyMRC is examples in dictionary, we carefully filter out the entries in test set from pre-training corpus, and only take the MEM and CL4SA as pre-training tasks. The results are shown in \Cref{tab:poly_res}. Compared to baselines, \textsc{CDBert} shows a $1.01\%$ improvement for accuracy on average. We notice that the overall performance show weak relation with the scale of the training corpus of PLM, which is a good sign as it reveals that the new task can not be solved by models simply adding training data.

\subsection{FewShot Setting on PolyMRC and CCLUE-MRC}

\begin{table}[th!]
    \centering
    \caption{Performance of \textsc{CDbert} on 10-shot setting of two MRC benchmarks.}
    \label{tab:cclue_10shot}
    \resizebox{\linewidth}{!}{%
    \begin{tabular}{l|ccccc}
    \thickhline
        \multicolumn{1}{c|}{\textbf{Model}}  & \textbf{PolyMRC} &\textbf{CCLUE-MRC}   \\ \hline
         BERT$_{\text{Base}}$ & 30.98 & 23.68 \\ 
         BERT$_{\text{Base}} + $ \textsc{CDBert}& 36.65 & 28.05 \\ 
         RoBERTa$_{\text{ext}}$ & 28.85 & 26.67 \\ 
         RoBERTa$_{\text{ext}} + $ \textsc{CDBert}& 29.47 & 28.51 \\ 
         RoBERTa$_{\text{ext-large}}$ & 28.45 & 25.06 \\ 
         RoBERTa$_{\text{ext-large}} + $ \textsc{CDBert}& 29.35 & 27.59 \\ 
         MacBERT$_{\text{ext-large}}$ & 37.35 & 25.29 \\ 
         MacBERT$_{\text{ext-large}} + $ \textsc{CDBert}& 39.22 & 27.81 \\ 
         
         \thickhline
    \end{tabular}
    }
\end{table}

To further investigate the ability of \textsc{CDBert} on few-shot setting, we construct two challenge datasets based on CCLUE MRC and PolyMRC. Following Few CLUE benchmark [few CLUE], we collect 10 samples for these two MRC tasks. Additionally, we build three different training samples to alleviate the possible fluctuating results of models training on small datasets. We demonstrate the results on \Cref{tab:cclue_10shot}. Compared with BERT, \textsc{CDBert}+BERT improves on accuracy from 30.98 to 36.65 ($18.3\%$ relatively) on PolyMRC, from 23.68 to 28.05 ($18.45\%$ relatively) on CCLUE-MRC. The performance gain on BERT is much more significant than larger baselines. This observation indicates that \textsc{CDBert} is promising in semantics understanding with a handful of annotated training data.

\subsection{Ablation Study}

\begin{table}[ht!]
    \centering
    \caption{Ablation of \textsc{CDBert} on CCLUE-MRC.}
    \label{tab:cclue_ablation}
    \begin{tabular}{l|c}
    \thickhline
        \multicolumn{1}{c|}{\textbf{Model}}  & \textbf{ACC}  \\ \hline
         RoBERTa+\textsc{CDBert} & 44.14 \\ 
         RoBERTa & 42.30 \\ 
         - Radical & 43.68 \\
         Replace with Glyph & 42.53 \\ 
         Replace with Char Dict & 42.76 \\ 
         w/o. CL4SA & 43.68 \\ 
         w/o. EL & 42.99 \\ 
         Continuous-pre-train & 43.14 \\
         \thickhline
    \end{tabular}
\end{table}

We conduct ablation studies on different components of \textsc{CDBert}. We use the CCLUE-MRC for analysis and take the Roberta$_{base}$ as the backbone. The overall results are shown in \Cref{tab:cclue_ablation}. Generally, \textsc{CDBert} improves the Roberta from $42.30$ to $44.14$ ($4.3\%$ relatively). 
\paragraph{The Effect of Character Structure}
We first evaluate the effects of radical embeddings and glyph embeddings. For fair comparisons, we keep other settings unchanged, and focus on the following setups: "-Radical", where radical embedding is not considered; "Rep Glyph", where we replace the radical embedding with glyph embedding. Results are shown in row 3-4. As can be seen, when we replace the radical embedding with glyph embedding, the accuracy drops 1.61 points, where the performance degradation is more obvious than removing radical embedding. The reason we use here is the scale of training corpus is not large enough to fuse the pre-trained glyph feature to \textsc{CDBert}. 

\paragraph{The Effect of Dictionary}
We then assess the effectiveness of the dictionary. We replace the original dictionary with character dictionary (row 5) and keep the model size and related hyper-parameters the same as \textsc{CDBert} pre-training procedure for fair. Besides, during finetuning process, we identify all the characters that are included in the character dictionary for further injecting with dictionary knowledge. We observe the character \textsc{CDBert} is helpful to some degree ($1.1\%$) but is much worse than the original \textsc{CDBert}. On the one hand, the number of characters in Chinese is limited, on the other hand, a word and its constituent characters may have totally different explanations. 

\paragraph{The Effect of Pre-training Tasks}
At last, we evaluate different pre-training tasks of \textsc{CDBert} including CL4SA and EL (row 6-7). Specifically, both CL4SA and EL help improve the NLU ability of PLM, and EL demonstrated larger improvement than CL4SA. The average improvements on CCLUE-MRC brought by CL4SA and EL are $1.05\%$ and $2.68\%$. In order to verify the impact of \textsc{CDBert} instead of the additional corpus, we follow \citet{Cui2019PreTrainingWW} to continuously pre-train the Roberta on the dictionary, which is regarded as extended data. As shown in row 8, using additional pre-training data results in further improvement. However, such improvement is less than our proposed \textsc{CDBert}, which is a drop of 1 point. 



\section{Limitations}
We collect the dictionary from the Internet, and although we make effort to reduce replicate explanations, there is noise in the dictionary. Besides, not all the words are included in the dictionary. In other words, the quality and amount of entries in the Chinese dictionary are to be improved. Additionally, our method is pre-trained on the Bert-like transformers to enhance the 
corresponding PLMs, and can not be applied to LLM directly whose frameworks are unavailable. In the future, we will use the retriever for disambiguation and dictionary knowledge infusion to LLM.

\section{Conclusion}
In this work, we leverage Chinese dictionary and structure information of Chinese characters to enhance the semantics understanding ability of PLM. To make Chinese Dictionary knowledge better act on PLM, we propose 3 pre-training objectives simulating looking up dictionary in our study, and incorporate radical or glyph features to \textsc{CDBert}. Experiment results on both modern Chinese tasks and ancient Chinese tasks show our method significantly improve the semantic understanding ability of various PLM. In the future, we will explore our method on more high-quality dictionaries~(\eg Bilingual dictionary), and adapt our method to LLM to lessen the semantic errors. Besides, we will probe more fine-grained structure information of logograms in both understanding and generation tasks.

\section*{Acknowledgements}
This project is supported by National Key R\&D Program of China (2021ZD0150200).

\bibliography{anthology,custom}
\bibliographystyle{acl_natbib}

\appendix






\end{document}